# A Hybrid Quantum-Classical AI-Based Detection Strategy for Generative Adversarial Network-Based Deepfake Attacks on an Autonomous Vehicle Traffic Sign Classification System

**M Sabbir Salek, Ph.D.**
Senior Engineer
National Center for Transportation Cybersecurity and Resiliency (TraCR)
414A One Research Dr, Greenville, SC 29607, USA
Email: msalek@clemson.edu

**Shaozhi Li, Ph.D.**
Post-doctoral Research Associate
National Center for Transportation Cybersecurity and Resiliency (TraCR)
414A One Research Dr, Greenville, SC 29607, USA
Email: shaozhl@clemson.edu

**Mashrur Chowdhury, Ph.D., P.E.**
Eugene Douglas Mays Chair of Transportation
Director, National Center for Transportation Cybersecurity and Resiliency (TraCR)
Glenn Department of Civil Engineering, Clemson University
216 Lowry Hall, S Palmetto Blvd, Clemson, SC 29634, USA
Email: mac@clemson.edu

Word Count: 5,323 words + 2 tables (250 words per table) = 5,823 words

*Submitted August 1, 2024*

**ABSTRACT**
The perception module in autonomous vehicles (AVs) relies heavily on deep learning-based models to detect and identify various objects in their surrounding environment. An AV traffic sign classification system is integral to this module, which helps AVs recognize roadway traffic signs. However, adversarial attacks, in which an attacker modifies or alters the image captured for traffic sign recognition, could lead an AV to misrecognize the traffic signs and cause hazardous consequences. Deepfake presents itself as a promising technology to be used for such adversarial attacks, in which a deepfake traffic sign would replace a real-world traffic sign image before the image is fed to the AV traffic sign classification system. In this study, the authors present how a generative adversarial network-based deepfake attack can be crafted to fool the AV traffic sign classification systems. The authors developed a deepfake traffic sign image detection strategy leveraging hybrid quantum-classical neural networks (NNs). This hybrid approach utilizes amplitude encoding to represent the features of an input traffic sign image using quantum states, which substantially reduces the memory requirement compared to its classical counterparts. The authors evaluated this hybrid deepfake detection approach along with several baseline classical convolutional NNs on real-world and deepfake traffic sign images. The results indicate that the hybrid quantum-classical NNs for deepfake detection could achieve similar or higher performance than the baseline classical convolutional NNs in most cases while requiring less than one-third of the memory required by the shallowest classical convolutional NN considered in this study.

**INTRODUCTION**
With worldwide leading industries, like Tesla, Waymo, General Motors, and Baidu, focusing heavily on autonomous vehicle (AV) development, the opportunity of riding a vehicle without ever needing to drive it is getting more realistic. An AV depends on its perception module to perceive the surrounding environment, to recognize different types of objects, such as other vehicles, pedestrians, and road users, and to realize roadway regulations enforced by traffic signals and signs (*1*). This perception module leverages different onboard sensors like camera, radio detection and ranging (radar), and light detection and ranging (lidar) to collect data from its surrounding environment. These data are processed and analyzed in the perception module to extract relevant information for autonomous driving.

Recognizing traffic signs, either regulatory or warning, posted on the roadway is crucial for driving safely through roadways. Failure to detect and identify these signs could lead to hazardous consequences, especially for AVs. An AV's perception module includes object detection and classification systems, such as a traffic sign recognition system, to carry out this task. However, recent studies have indicated that such systems could be attacked by tampering with the captured images or the classification models, causing the perception system to misrecognize the traffic signs (*2*). These attacks are commonly known as adversarial attacks and have been widely studied in recent years, where researchers have introduced various adversarial attack models and their corresponding detection, mitigation, and resilient strategies.

With the recent advent of deep learning, producing realistic fake images and videos has become common. Deep learning-based techniques for manipulating existing images/videos are known as deepfake (*3*, *4*). Anyone with a computer or a smartphone can manipulate or create realistic fake images or videos nowadays and cause the spread of misinformation. Although, to the authors' best knowledge, no existing studies have considered a deepfake technique as a means to perform an adversarial attack on the AV perception module, it is quite feasible to conduct such attacks.

In this study, the authors explored how a deepfake technique can be utilized to perform an adversarial attack on an AV traffic sign classification system, leading the system to misrecognize any traffic sign images. The authors used a generative adversarial network (GAN) based approach to perform the deepfake attacks in this study. The deepfake attack was performed on real-world traffic sign images from two benchmark datasets. Furthermore, the authors introduced an amplitude encoding-based hybrid quantum-classical neural network (NN) supported strategy to detect the deepfake attack-generated fake traffic sign images. Several classical deep learning models, ranging from a shallow two-layer convolutional NN (CNN) to a six-layer deep CNN, were employed to serve as baseline models to compare with the deepfake detection performance achieved by the hybrid quantum-classical NN-based strategy. Our results indicated that the hybrid strategy could achieve a detection performance comparable to that of the classical strategy. However, due to quantum phenomena such as quantum superposition and entanglement and the adoption of amplitude encoding for transforming image data into representative quantum states, the memory requirement for the hybrid quantum-classical NN models was found to be much lower than the classical CNN models.

The contributions of this study are as follows: (i) the development of a GAN-based deepfake attack model to perform adversarial attacks on an AV traffic sign classification system, and (ii) the development of a hybrid quantum-classical NN-based deepfake detection strategy that can achieve similar detection performance to classical CNN models with substantially lower memory requirement. The rest of the paper is organized as follows: the second section summarizes some of the recent noteworthy studies conducted for deepfake detection; the third section presents the deepfake attack model and the detection strategy developed in this study; the fourth section presents the dataset and the evaluation metrics used to evaluate the performance of the deepfake detection strategy introduced here, followed by the evaluation results and related discussions; and finally the last section highlights the conclusions of this study.

**LITERATURE REVIEW**
In this section, the authors present a brief overview of the recent advances in deepfake detection technology. The existing research on deepfake technology primarily focuses on human images or videos.

The first deepfake image was created to replace a person's face in pornographic videos using GANs in 2017. After that, many progress has been made in generating deepfake images, including Face2Face (a computer graphics technique-based face animation software) (*5*) and CycleGAN (*6*). Rana et al. (*7*) performed a systematic literature review and classified the existing deepfake detection techniques into four classes: (i) machine learning-based methods (*8*), (ii) deep learning-based methods (*9*), (iii) statistical measurement-based methods (*10*), and (iv) blockchain-based methods (*11*). Among these, the deep learning-based method is the most widely used technique due to its higher accuracy compared to the other methods. The deep learning-based method with Meso-4 module was first proposed for deepfake video detection using images from the Face2Face dataset by Afchar et al. in 2018 (*12*). Later, the detection technique was improved by using features extracted from other neural networks, such as the handcrafted features (*13*), spatiotemporal features (*14*, *15*), common textures (*16*), and face landmarks with visual artifacts (for example, eye, teeth, and lip movement) (*17*). Recently, a capsule network with an attention mechanism was proposed, which needs a smaller number of parameters. This method was evaluated on the FaceForensics++ dataset and achieved more than 99% accuracy (*18*).

Motivated by the advances in quantum computing and machine learning, quantum NN has been applied for deepfake detection using the celeb deepfake dataset by Mishra and Samanta (*19*). The authors of (*19*) used a pre-trained ResNet-18 model to extract features that were fed into a quantum NN to detect deepfake images and obtained an accuracy of 96.1%. However, this method lacks efficiency as the ResNet-18 model has to be trained for feature extraction before applying the quantum NNs.

The existing body of literature on deepfake technology does not consider its applicability and corresponding threats to the transportation industry, in particular for autonomous driving-related tasks that involve image recognition. To this end, in this study, the authors introduced a deepfake attack model that can be used to perform an adversarial attack on an AV traffic sign classification system. The authors developed a hybrid quantum-classical NN-based strategy to detect deepfake traffic sign images. Unlike the previous study based on quantum machine learning (*19*) that did not consider quantum NNs for feature extraction, the hybrid strategy presented here leverages quantum NNs to extract features and a fully connected classical neural network to do the classification. In addition, unlike the approach presented in (*19*), the quantum NNs and the classical fully connected layers of this hybrid NN model can be trained simultaneously, yielding better efficiency in training.

**DEEPFAKE ATTACK AND DETECTION METHOD**
In this section, the authors introduce the deepfake attack model, which aims to fool an AV traffic sign recognition system and cause it to misrecognize the traffic signs it comes across on the roadway. Next, the authors present a deepfake detection strategy, including a classical CNN-based detection strategy used as the baseline and a hybrid quantum-classical NN-based deepfake detection strategy with reduced memory requirements.

**Deepfake Attack Model**
An AV traffic sign recognition system receives roadway images captured by the onboard camera(s), detects and then classifies the traffic signs in them to help the AV perceive its driving environment, comply with roadway regulations, take actions based on warnings, and navigate safely through its desired route. Similar to any other adversarial attacks, the authors assume that an attacker resides between the onboard camera-captured traffic sign image and the AV's traffic sign classification system and is capable of manipulating the image data as the attacker wishes (as shown in **Figure 1**). Upon receiving a traffic sign image, the attacker leverages a deepfake attack module to instantly generate a fake traffic sign image that is different from the input traffic sign type and replaces the real image with the fake one.

As depicted in **Figure 1**, the deepfake module contains a pretrained generator that generates a fake traffic sign image for an input image $x$ (i.e., a real traffic sign image). This is done by optimizing the input latent vector $z$, which is the only input to the generator. The optimal input latent vector $z^*$ for generating the fake image that would replace the real one is obtained by solving the optimization problem presented in **Equation 1** using a gradient descent-based iterative approach,

$$z^* = \arg\max_{z} \|G(z) - x\|_2^2 \tag{1}$$

where, $G(z)$ presents the generated image for the input latent vector $z$. The optimization ensures that the fake image $G(z^*)$ that will replace the real one least resembles the input image $x$; therefore, it is most likely that the fake image $G(z^*)$ would be of a different type of traffic sign from the input image $x$. Because of the non-convex nature of the objective function in **Equation 1** and to ensure that the attack is feasible in real-time, the authors assume that the attacker utilizes a fixed number of random initializations of $z$.

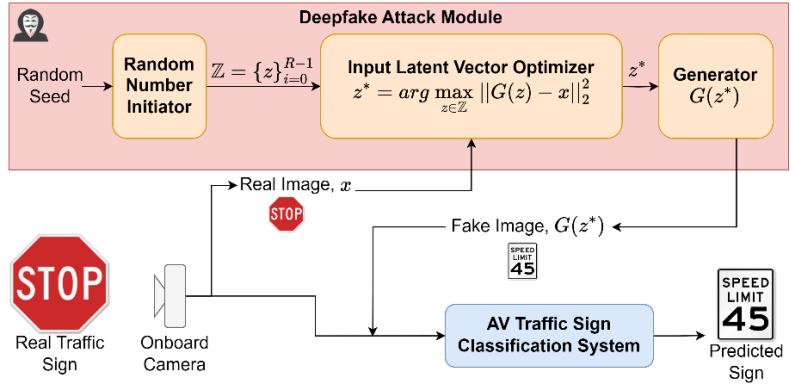

**Figure 1 Deepfake Attack Model**

The generator model is obtained through training a GAN model using the Wasserstein GAN with gradient penalty (WGAN-GP) approach introduced by Gulrajani et al. (*20*). Like any other GAN models, WGAN-GP includes a generator (to produce fake images) and a discriminator (to tell apart the generator-produced fake images from the real ones) that are trained in tandem using a min-max loss function given by **Equation 2** below.

$$\min_{G}\max_{D} V_W(D,G) = \mathbb{E}_{x\sim P_r(x)}[\log D(x)] - \mathbb{E}_{z\sim P_f(z)}\left[\log\left(D\big(G(z)\big)\right)\right] + \lambda \mathbb{E}_{\hat{x}\sim P_{\hat{x}}}[\|\nabla_{\hat{x}} D(\hat{x})\|_2 - 1]^2 \tag{2}$$

where, $x$ and $z$ are the input image and the input latent vector, respectively; $G(\cdot)$ and $D(\cdot)$ represent the generator and the discriminator, respectively; $E(\cdot)$ represents the expected value; $P_r(\cdot)$ and $P_f(\cdot)$ represent the distributions of real and generated (fake) images, respectively; $\lambda$ is a constant, which is set to 10 (*20*); $\nabla_{\hat{x}}$ is a differential operator with respect to $\hat{x}$; and $\hat{x}$ is sampled from $x$ and $G(x)$ using **Equation 3** given below.

$$\hat{x} = tG(x) + (1-t)x \tag{3}$$

where, $t$ is uniformly sampled between 0 and 1, i.e., $0 \leq t \leq 1$.

The neural network architectures used for the generator and the discriminator are based on the deep convolutional GAN (DCGAN) (*21*) and WGAN (*20*), as shown in **Figures 2** and **3**, respectively. In **Figures 2** and **3**, each layer's output dimension is shown as $C \times H \times W$, where $C$, $H$, and $W$ represent the number of feature maps, and the height and the width of each feature map, respectively.

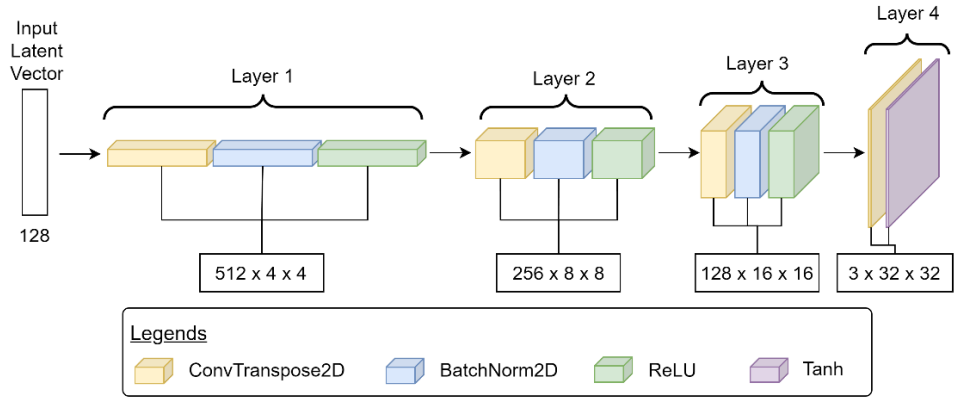

**Figure 2 Generator Architecture (Based on DCGAN)**

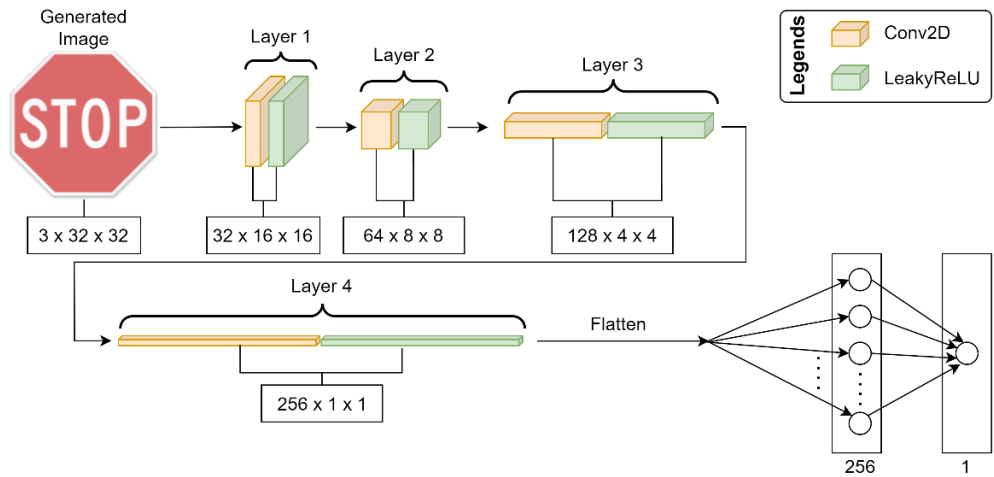

**Figure 3 Discriminator Architecture (Based on WGAN)**

**Deepfake Detection Strategy**
In this subsection, the authors present the deepfake detection strategy developed in this study. This strategy employs two NNs, as shown in **Figure 4**. The first NN, a CNN (i.e., a traffic sign classifier), is responsible for recognizing the traffic sign type of an input traffic sign image. The authors utilize a ResNet9 (*22*) based architecture (as shown in **Figure 5**) to develop the traffic sign classifier. After the traffic sign type is determined by the traffic sign classifier, the second NN (i.e., a deepfake detection model) determines whether the image is a real image or a fake image. The deepfake detection model is a traffic sign type-specific model, i.e., this strategy requires a pretrained deepfake detection model for each type of traffic sign. For developing the traffic sign type-specific deepfake detection models, the authors take two approaches: (i) classical CNN-based deepfake detection and (ii) quantum-classical NN-based deepfake detection; the following subsections present these approaches in detail.

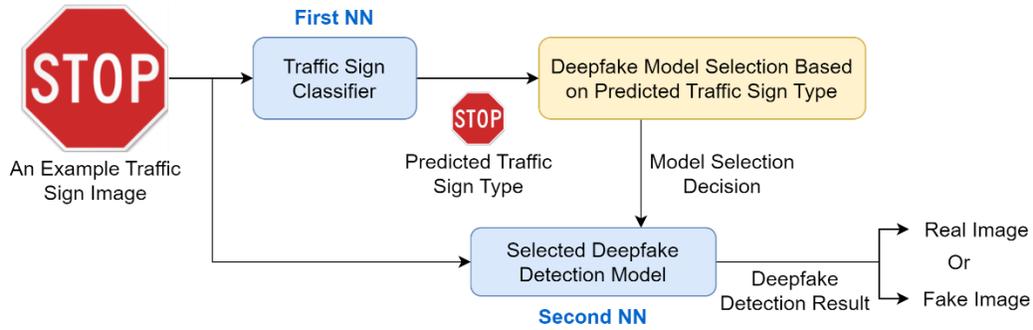

**Figure 4 Deepfake Detection Strategy**

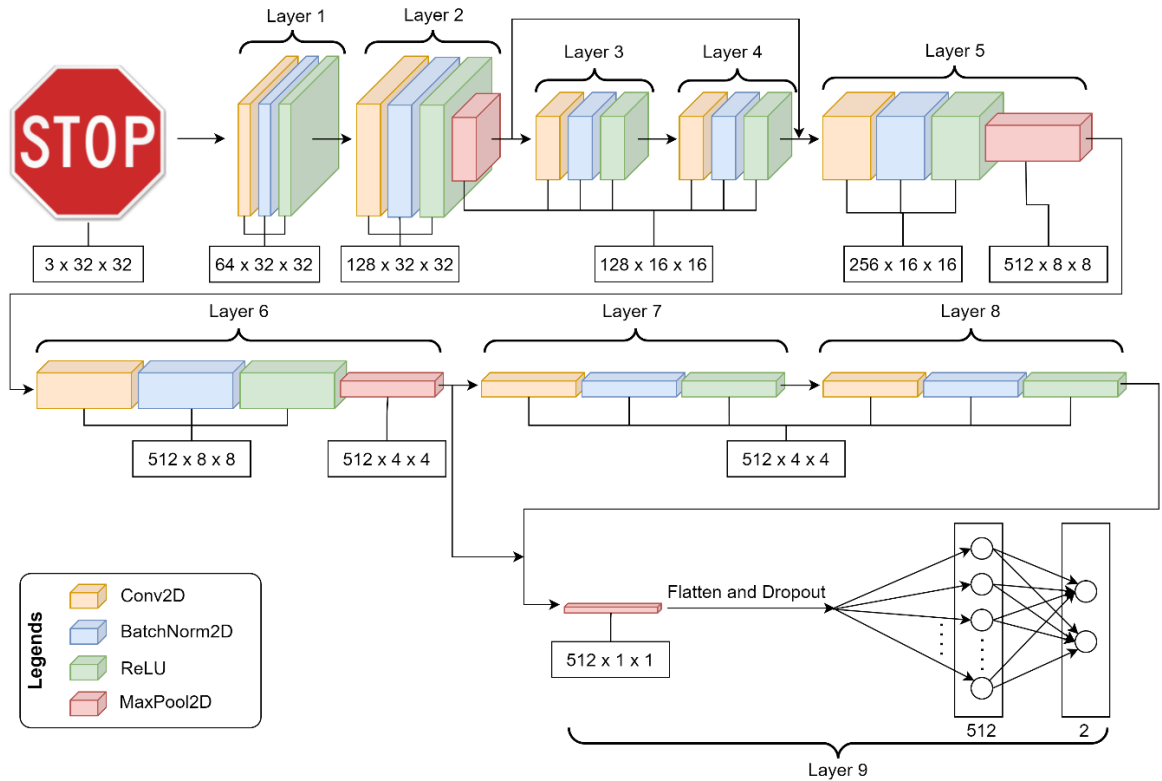

**Figure 5 Traffic Sign Classifier Architecture (Based on ResNet9)**

*Classical CNN-based Deepfake Detection*
The authors consider five different classical CNN models to have a set of baseline classical deepfake detection models. These models vary from a shallow two-layer CNN model to a deeper six-layer CNN model as follows: (i) Classical CNN-1 (includes one convolutional layer and one fully connected layer), (ii) Classical CNN-2 (includes two convolutional layers and one fully connected layer), (iii) Classical CNN-3 (includes three convolutional layers and one fully connected layer), (iv) Classical CNN-4 (includes four convolutional layers and one fully connected layer), and (v) Classical CNN-1 (includes five convolutional layers and one fully connected layer). **Figure 6** presents the architectures for the first three classical CNNs used in this study as examples. The remaining two classical CNNs, i.e., classical CNN-4 and CNN-5, also follow the same architectures with one and two additional convolutional layers, respectively. Note that the shape boxes in **Figure 6** are not drawn to scale.

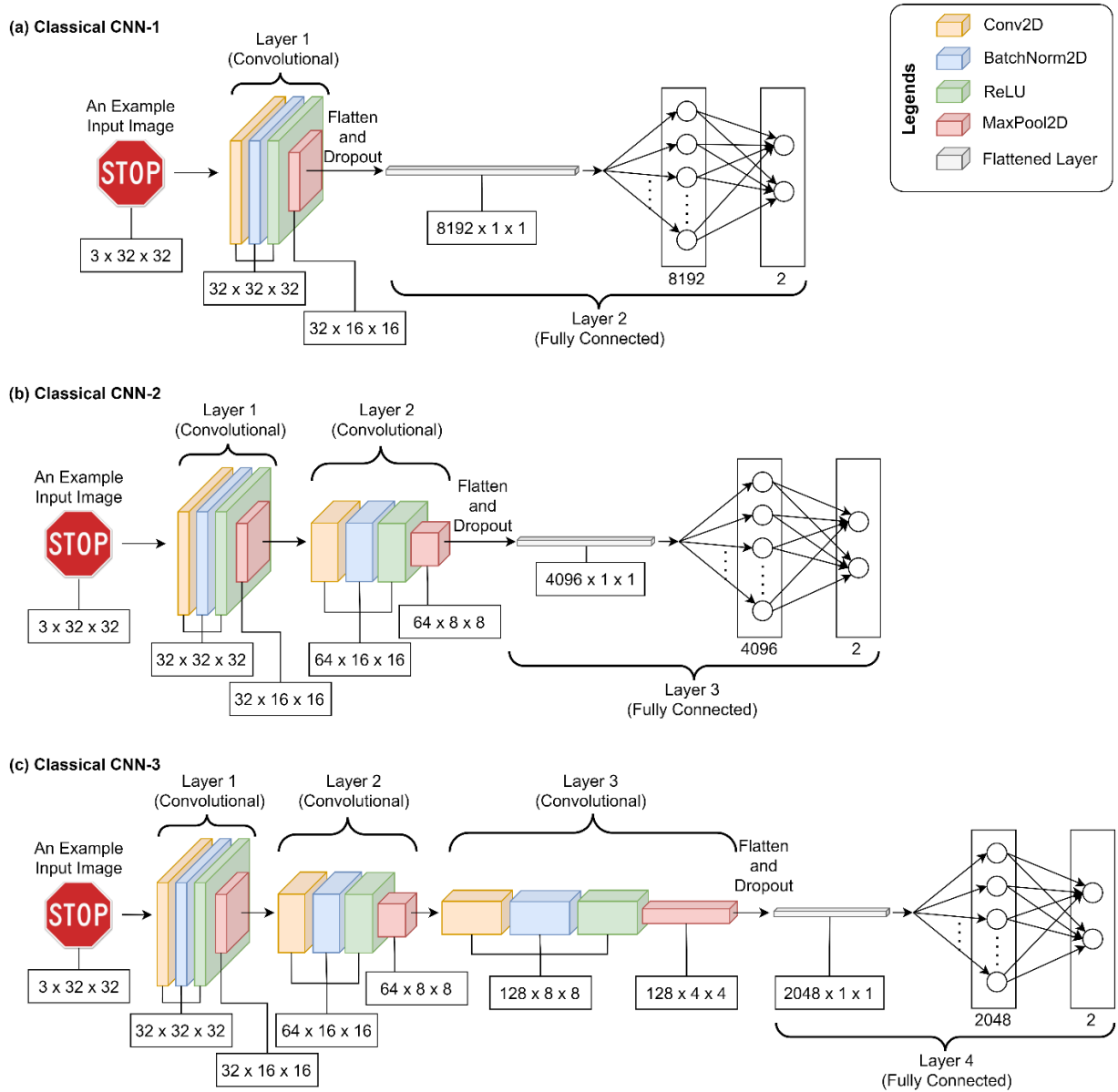

**Figure 6 The Architectures of (a) the Classical CNN-1, (b) the Classical CNN-2, and (c) the Classical CNN-3.**

As observed from **Figure 6**, each convolutional layer in the classical CNNs includes a convolutional step (denoted as Conv2D) with a $3 \times 3$ kernel and padding (padding size: 1) applied, followed by a batch normalization step (denoted as BatchNorm2D), a rectified linear unit-based activation step (denoted as ReLU), and a max pooling step (denoted as MaxPool2D). For each of the classical CNNs, the output feature map from the last convolutional layer is flattened, and a 20% dropout of the neurons is applied as a regularization to reduce overfitting issues (*23*) before feeding the feature map to the fully connected layer. The final output of the classical CNNs maps to two labels: real image (labeled as 0) and fake image (labeled as 1). The authors implemented the classical CNNs using the open-source PyTorch library (*24*) in Python.

*Hybrid Quantum-Classical NN-based Deepfake Detection*
The authors leverage a hybrid quantum-classical NN to detect fake images for each of the traffic sign types. The hybrid NN includes quantum NNs and two fully connected layers, as shown in **Figure 7(a)**. The authors utilized 32 quantum circuits to generate 32 neurons in the first hidden layer, which is connected to the second hidden layer with 120 neurons. The activation function between the first and the second hidden layers is set as $Tanh$.

We transform the data of an input traffic sign image to a quantum state using amplitude encoding. Therefore, the initial quantum state is set as,

$$|\psi_0\rangle = \sum_i \frac{x_i}{|x|} |b_i\rangle \quad (4)$$

where, $x_i$ denotes the color intensity at the $i$-th site in the image and $b_i$ is the basis, denoting the binary representation of the number $i$. For a $3 \times 32 \times 32$ image with 3,072 pixels, 12 qubits are used (it has $2^{12}$ basis) to represent it. Compared to the frequently used angle-encoding method (*25*), amplitude encoding can take advantage of the large Hilbert space and the powerful storage capability of quantum computers. As shown in **Figure 7(b)**, each quantum NN consists of one rotational layer, three quantum convolutional layers, and three quantum pooling layers. The rotational layer is made of single qubit rotation operators $R_y(\theta_1)$ and $R_z(\theta_2)$. The two-gate operators in the quantum convolutional layer (denoted as $U(\varphi)$) and the quantum pooling layer (denoted as $P(\varphi)$) are presented in **Figures 7(c)** and **7(d)**, respectively. The output of the quantum circuit is the average value of $O = \langle\psi|Z_{12}|\psi\rangle$, where $|\psi\rangle$ is the final quantum state and $Z_{12}$ denotes the z-component of the Pauli matrix on the 12-th qubit.

(a) Hybrid quantum-classical NN

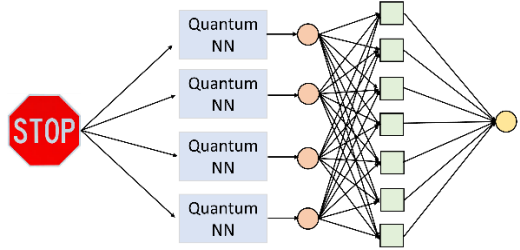

(b) Quantum NN

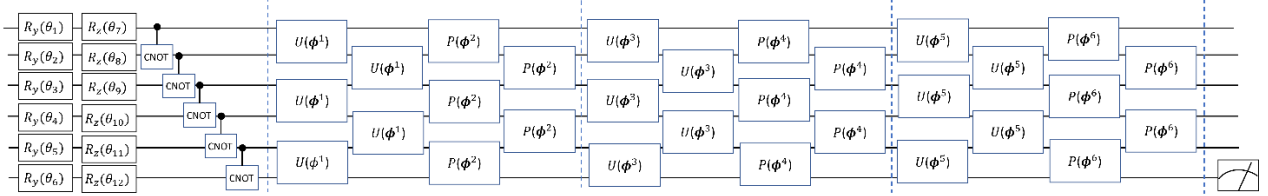

(c) $U(\phi)$

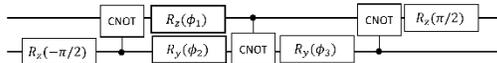

(d) $P(\phi)$

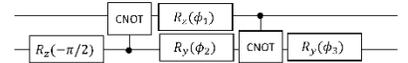

**Figure 7 The Architectures of (a) the Hybrid Quantum-Classical NN, (b) the Quantum NN, (c) the Two-Gate Operator $U(\varphi)$, and (d) the Two-Gate Operator $P(\varphi)$ Used in This Study.**

The authors used the Adam optimizer to optimize parameters in the hybrid quantum-classical NN. In principle, the gradient of $O$ with respect to the rotation angle can be exactly evaluated using the parameter shift rule, which needs to simulate a quantum circuit three times. To reduce the computational cost, we evaluate this gradient using the finite difference method. The authors used the open-source

Python library, PennyLane (*26*), to simulate the quantum NNs. Unlike a real quantum computer, the quantum simulator does not have noises, making the finite difference method for gradient computations reliable.

**EVALUATION**
This section explains the evaluation approach considered in this study. First, the authors present the details of the traffic sign dataset and the deepfake dataset used here. Second, the authors introduce the evaluation metrics considered for evaluating the performance of our classical and hybrid quantum-classical deepfake detection models.

**Dataset**
The deepfake detection strategy presented in this study requires adversarial training. Therefore, the dataset used to train the deepfake detection models must have both real traffic sign images (captured during driving in the real world) and fake traffic sign images (generated by the deepfake attack model introduced in this study). In this section, the authors present the details of the dataset considered in this study.

*Real-World Traffic Sign Image Dataset*
Since the deepfake attack model developed in this study relies on a GAN-based framework, which requires a lot of training data to produce realistic fake images, the authors reviewed the existing datasets containing US traffic sign images. Considering the unavailability of a single dataset containing enough images for each of the US traffic sign types, the authors decided to combine data from two prominent traffic sign image datasets, i.e., the LISA traffic sign dataset (*27*) and the Mapillary traffic sign dataset (*28*).

The LISA dataset comprises 49 US traffic signs types with 7,855 image annotations. These traffic sign images were extracted from video frames recorded by dashboard cameras of multiple vehicles traveling around San Diego, California. On the other hand, the Mapillary dataset comprises 313 types of worldwide traffic signs (out of which the authors only considered the US traffic sign images) and includes a total number of 82,724 traffic sign images. After combining the two datasets, the authors selected the ten types of traffic signs with the maximum number of available images and manually cleaned the dataset to ensure the final dataset of real traffic sign images only includes those recognizable by humans.

*Deepfake Traffic Sign Image Dataset*
The real traffic sign image dataset was used to train a GAN model to generate realistic fake images. After the GAN is trained, the authors used the images from the real traffic sign image dataset as inputs to the generator model to generate random fake traffic sign images based on the deepfake attack model presented in the earlier section. Even though the GAN model was adequately trained to generate realistic fake images, it still produces a few ambiguous images that do not fall under any particular type of traffic sign. This is due to the fixed number of iterations used in the deepfake attack model to generate a fake traffic sign image in real time that most closely resembles a traffic sign type other than the type of the input traffic sign image. These unrecognizable fake images were manually removed from the pool of deepfake attack-generated images to obtain the final dataset of fake traffic sign images.

The real and the fake images from each of the traffic sign types were then combined by randomly undersampling the majority class between the real and the fake images. This ensured that for each type of traffic sign, the numbers of real and fake images were balanced (i.e., 50% real images and 50% fake images) in the dataset. The total number of images for each traffic sign type is as follows: STOP (154 images), SIGNAL AHEAD (250 images), PED XING (194 images), SPEED LIMIT 25 (192 images), SPEED LIMIT 30 (190 images), SPEED LIMIT 35 (250 images), SPEED LIMIT 45 (200 images), SPEED LIMIT 55 (146 images), and STOP AHEAD (202 images). **Figure 8** presents a few examples of the real and fake images of each traffic sign type. Images from each traffic sign type were then randomly shuffled and divided into three subsets: training set (contains 60% of the images), validation set (contains 20% of the images), and test set (contains the remaining 20% of the images).

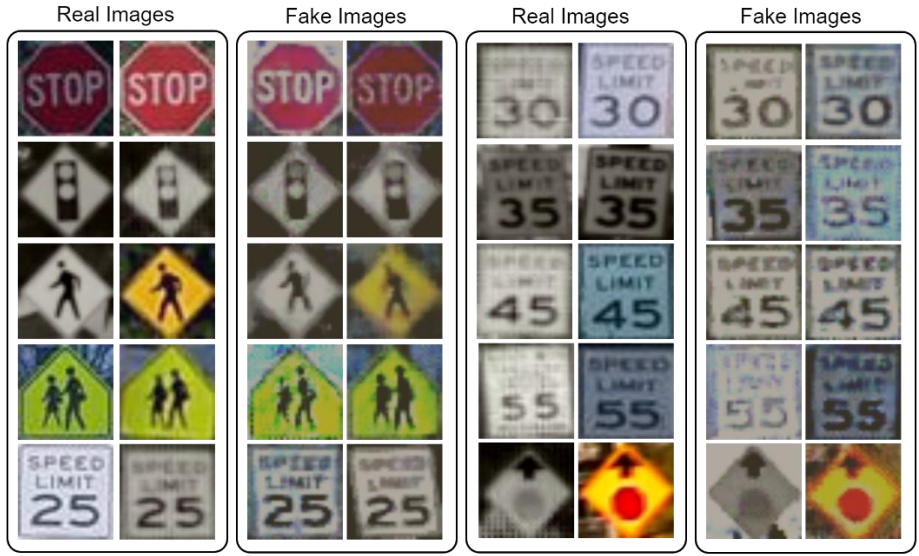

**Figure 8 Examples of Real and Fake Images from the Dataset Used in This Study**

**Evaluation Metrics**
Both two CNNs of the deepfake detection strategy introduced here, as previously shown in **Figure 1**, are tasked with classification problems. Therefore, the authors selected the four most popular evaluation metrics for deep learning-based classification problems: (i) precision, (ii) recall, (iii) F1 score, and (iv) accuracy to evaluate the performance of the CNN models. **Table 1** presents how these metrics were calculated and what they represent.

**TABLE 1 Evaluation Metrics**

| Evaluation Metric | How Metric is Calculated | Meaning |
|---|---|---|
| Precision | $\dfrac{True\ Positives}{True\ Positives + False\ Positives}$ | • Precision measures the ratio of correctly identified positives to all the predicted positives.<br>• A high precision implies low false positive errors. |
| Recall | $\dfrac{True\ Positives}{True\ Positives + False\ Negatives}$ | • Recall measures the ratio of correctly identified positives to all the actual positives.<br>• A high recall implies low false negative errors. |
| F1 Score | $2 \times \dfrac{Precision \times Recall}{Precision + Recall}$ | • F1 score is the harmonic mean of precision and recall, representing a single metric that balances both concerns.<br>• A high F1 score indicates that both precision and recall are high and well-balanced. |

*Salek, Li, and Chowdhury*

| Evaluation Metric | How Metric is Calculated | Meaning |
|---|---|---|
| Accuracy | $\frac{True\ Positives + True\ Negatives}{Total\ Number\ of\ Predictions}$ | • Accuracy is the proportion of the correct predictions among all the predictions made.<br>• A high accuracy indicates that the overall effectiveness of the model is high. |

In addition, to estimate the memory requirements for the deepfake detection models, the authors determine the total number of trainable parameters in each NN model. The trainable parameters for a NN include the weights and the biases of the NN model. Thus, the total number of trainable parameters in a NN can be determined using **Equation 5** as follows,

$$N_{total} = N_i \times N_{h_1} + \sum_{i=2}^{n-1}(N_{h_{i-1}} \times N_{h_i}) + N_{h_n} \times N_o + \sum_{i=1}^{n} N_{h_i} + N_o \tag{5}$$

where, $N_{total}$ is the total number of trainable parameters, $N_i$ is the number of neurons in the input layer, $N_{h_i}$ is the number of neurons in the $i^{th}$ hidden layer, $N_o$ is the number of neurons in the output layer, and $n$ is the number of hidden layers. If a parameter is saved as a single-precision floating-point number, it requires 32 bits or 4 bytes of memory. The authors estimate the total size of memory needed to store an NN model in megabytes (MB) using **Equation 6** as follows,

$$Estimated\ Memory\ (MB) = 4 \times \frac{N_{total}}{1,048,576} \tag{6}$$

**Evaluation Results and Discussions**

In this section, the authors present the evaluation results of the deepfake detection strategy introduced in this study. As shown in **Figure 1**, the first CNN identifies the type of traffic sign present in an input image. The ResNet9-based traffic sign classifier, used as the first CNN in this study, achieved an overall precision, recall, F1 score, and accuracy of 0.97, 0.97, 0.97, and 0.97, respectively, on the combined dataset of real and fake images of all traffic sign types. After the traffic sign type is identified, the respective deepfake detection CNN model is invoked by the deepfake detection system to determine whether the image is a real image or a fake image.

**Figures 9(a)** through **9(d)** present a comparison among the baseline classical CNN-based deepfake detection models and the hybrid quantum-classical NN-based deepfake detection model in terms of precision, recall, F1 score, and accuracy, respectively. As observed in **Figure 9**, from the classical CNN-1 through CNN-5, the detection performance improved gradually for all the traffic sign types with a few exceptions. This is expected as the additional convolutional layers help improve prediction performance by extracting relevant features. The hybrid quantum-classical NN model outperformed, or at least provided, similar performance to the classical CNN-1 through CNN-5 in most cases, including a few exceptions, such as for the PED XING and the SPEED LIMIT 45 signs.

However, the advantage of the hybrid quantum-classical NN model over the classical CNN models becomes apparent when their memory requirements are compared. **Table 2** presents the total number of parameters (based on **Equation 5**) and the required memory size in MB (estimated using **Equation 6**) for all these deepfake detection models. The hybrid quantum-classical NN model used in this study requires only 5,425 parameters or 0.021 MB of memory, which is less than one-third of the parameters or memory needed for the shallowest baseline classical model used in this study, i.e., classical CNN-1. Except for the PED XING and the SPEED LIMIT 45 signs, the hybrid deepfake detection approach outperformed the classical approach for classical CNN-1 through 3 (as observed in Figure 9) even though the required number of parameters for the hybrid approach is substantially less than these three classical CNN models.

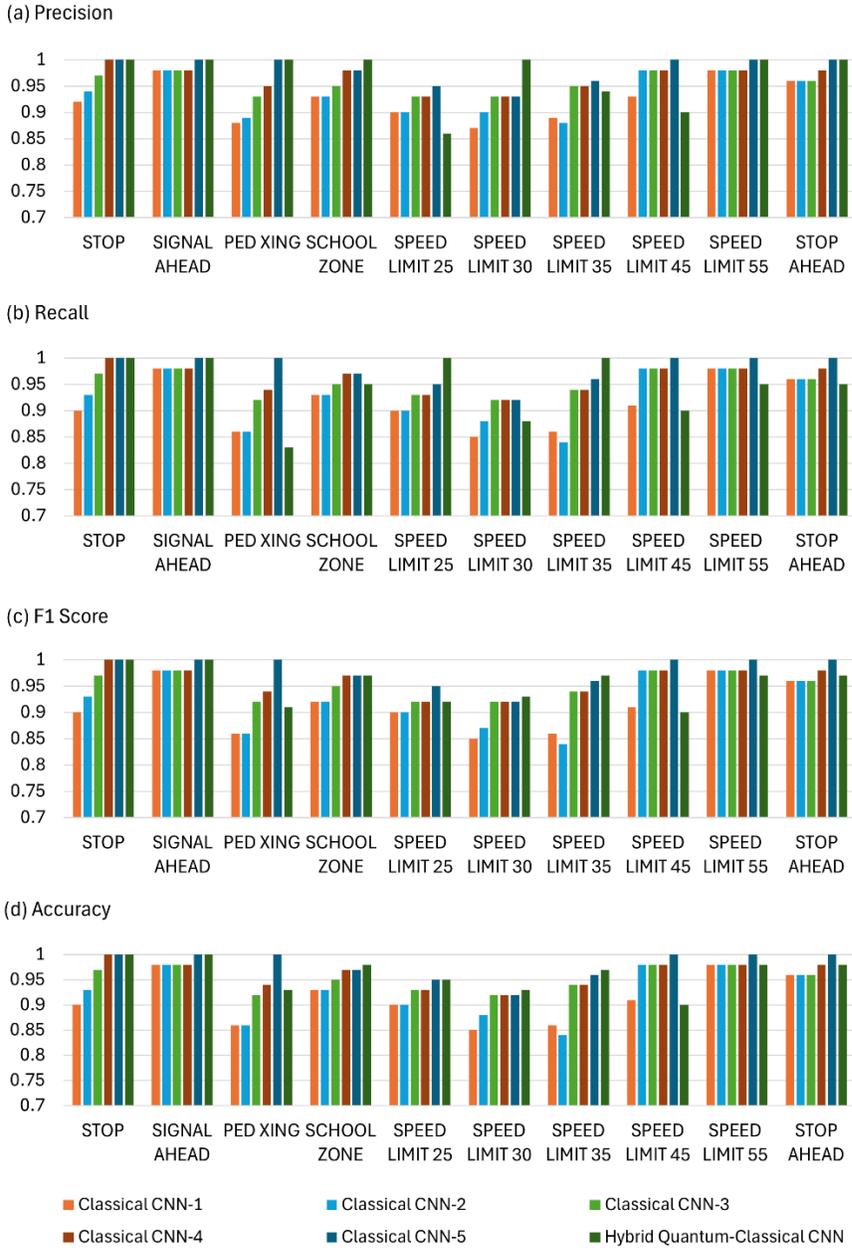

**Figure 9 Deepfake Detection Performance in Terms of (a) Precision, (b) Recall, (c) F1 Score, and (d) Accuracy**

**Table 2 Total Number of Model Parameters and Required Memory Size**

| Deepfake Detection Model | Total Number of Parameters | Estimated Memory Size |
|---|---|---|
| Classical CNN-1 | 17,346 | 0.066 MB |
| Classical CNN-2 | 27,778 | 0.106 MB |
| Classical CNN-3 | 97,794 | 0.373 MB |
| Classical CNN-4 | 391,426 | 1.493 MB |
| Classical CNN-5 | 1,571,586 | 5.995 MB |
| **Hybrid Quantum-Classical NN** | **5,425** | **0.021 MB** |

**CONCLUSIONS**
AVs rely on their onboard computers for a wide range of tasks, such as perception, localization, mapping, path planning, and motion control (*1*), that require large storage and processing capacity. Therefore, reducing the memory requirement for different autonomous driving-related tasks is crucial. In addition, the rapid expansion of attack surfaces due to the increased connectivity and automation warrants AVs to incorporate various cyberattack detection, mitigation, and resilient strategies posing an additional hurdle to the already existing challenge of memory allocation for AV onboard computers. Keeping this in mind, the authors developed a hybrid quantum-classical NN-based strategy for deepfake detection in this study.

Although deepfake-related security concerns have been explored extensively by researchers in recent years, their applicability to the transportation domain has not received any attention. This study introduced how this security threat affects our transportation systems by crafting a deepfake-based adversarial attack model aimed at fooling an AV perception module with fake traffic sign images. The hybrid quantum-classical NN-based deepfake detection strategy developed in this study could help detect such attacks effectively and efficiently without causing excessive load on the AV onboard computers. This strategy, evaluated on a real-world traffic sign image dataset combined with a deepfake traffic sign image dataset, obtained comparable deepfake detection performance to that of the baseline classical CNNs with a substantially low memory requirement for the model parameters.

A limitation of this study is that the authors could not develop deepfake detection models for all the US traffic sign types due to the limited availability of good-quality traffic sign images. In the future, the authors aim to expand the deepfake detection strategy introduced in this study to include all US traffic sign types.

**ACKNOWLEDGMENTS**
This work is based upon the work supported by the National Center for Transportation Cybersecurity and Resiliency (TraCR) (a US Department of Transportation National University Transportation Center) headquartered at Clemson University, Clemson, South Carolina, USA. Any opinions, findings, conclusions, and recommendations expressed in this material are those of the author(s) and do not necessarily reflect the views of TraCR, and the US Government assumes no liability for the contents or use thereof.

**AUTHOR CONTRIBUTIONS**
The authors confirm their contribution to the paper as follows: study conception and design: M.S. Salek, S. Li, and M. Chowdhury; data collection: M.S. Salek; analysis and interpretation of results: M.S. Salek and S. Li; draft manuscript preparation: M.S. Salek, S. Li, and M. Chowdhury. All authors reviewed the results and approved the final version of the manuscript.